# Logical AI OS: A Training-Free Governance Layer Outside the Model for Instruction Following — Benefits, Costs, Failures, and the Calibration of the Measuring Instrument Itself

**Joe Yuan** (ORCID 0009-0009-2803-8987), independent researcher.  

## Abstract

Users give language models explicit requests (length, number of items, format, paragraph structure, required or forbidden words, grammatical person, answer language, and so on), and small models often fail to keep them. This report describes a governance layer that sits outside the model and involves no training: it reads the user's requests with rules, checks the answer mechanically, repairs the answer when a request is not kept, and hands back the original answer when the repair fails. The loop itself exists in prior work. What this report does is measure it: its benefits, its costs, where it fails, and how the measuring instrument itself was calibrated.

On four Qwen-family models (3B, 7B, 14B and one 7B reasoning model, all 4-bit quantised) and 329 Chinese prompts: (1) Specification level — with the governance layer, the number of answers that keep every explicit request the layer read rises significantly (all six comparisons $p<0.0001$): by 58 to 88 answers under the original checker, 34 to 66 under the corrected checker, and 52 to 92 after the length control was corrected and the arms were rerun; the one generic redo message we tested, which does not say which request failed, brought no significant gain. (2) User level (exploratory) — on 90 prompts where the bare model failed and a repair was triggered, cases where the governance layer produced an answer the annotator could accept and the bare model did not outnumber the reverse (49 to 7 and 18 to 12 for the two AI annotators, the latter not significant; 43 to 3 and 19 to 0 after applying the rule "an answer short of the required length is not acceptable"). (3) Costs — 43% to 66% more generation time per answer; whole-answer rewriting damages content, and a sizeable share of originally usable answers become unusable. (4) Failures — rule-based reading has a limit on unseen phrasings (on the two most recent exams, the share of newly phrased requests it read was about 70% and 84% of the two-annotator ceiling; trap sentences were correctly left alone 85% of the time); mutually exclusive or unsatisfiable requests make the reasoning model loop in its thinking segment until the budget runs out; the model's statements about what it did are unreliable (in one 7B repair experiment, 18 of the 25 items it reported as "fixed" were not fixed). (5) The instrument — on requests that require parsing the structure of the answer, more than half of the original checker's "violated" verdicts were wrong; after correction and freezing the main conclusions hold, but one comparison (local patch better than whole-answer rewrite) is downgraded from "holds" to "same direction, unstable significance".

The conclusions hold only under these conditions: this set of Chinese material, one model family (Qwen), and model sizes up to 14B.

## 1 Introduction

**The problem.** When users ask a model to write something, they often attach explicit requests: "no more than 300 characters", "list five points", "answer in English", "do not use a certain word". Such requests share one property: there is a stated standard of right and wrong, and each instance can be checked on its own. Small models fail these requests at a non-trivial rate, and (as Section 7 shows) their statements about whether they kept them are unreliable.

**What this report does.** It studies a governance layer placed outside the model: read the requests, check, repair, deliver. This loop is not proposed here (see Section 2). The work here is to treat it as an object to be measured: how many answers it moves from not kept to kept; how much time it costs and how many originally usable answers it breaks; under what conditions it fails; and whether the instrument used to measure it is itself accurate, and which conclusions changed after the instrument was corrected.

The purpose of this research is the usability of AI for people: whether what the user asked for out loud was done, and whether the user is told truthfully when it was not.

**Three limitations to know before reading.**

1. **Reading has a limit.** The governance layer reads the user's requests with rules. On unseen phrasings it does not read all of them (Section 7.1). Every "kept" number in this report is measured within "the checkable requests the governance layer read": a request that was not read is outside the scoring, and a misread request is scored as if it were a request. The bare model and every governed arm use the same item list. The checker and the reading rules share no code, but what the checker checks is still the output of the reading step; this dependence is not removed in this report.
2. **Only one model family was tested.** All four models are in the Qwen family; the reasoning model is also built on a Qwen base. We have no data on other families or larger models.
3. **The specification level is not the user level.** The main numbers measure "was the specification followed", which is not the same as "did the user find it useful". The first comparison of the two is in Section 5 and is exploratory.

This report does not discuss the overall theoretical blueprint of the project.

## 2 Related work

The loop "check with an external tool after generation, and rewrite with feedback when the check fails" appears in several earlier papers. We claim no novelty for it.

**Closest prior work.** DVR (Zhang et al., Findings of ACL 2025) is training-free: the model splits the instruction into single constraints and selects tools; the tools check and return textual feedback with numbers; the model rewrites accordingly. CRITIC (Gou et al., ICLR 2024) lets a model correct itself with feedback from external tools. NSVIF (Su et al., arXiv:2601.17789) formalises "was the instruction followed" as a constraint-satisfaction problem and points out that a model as judge is unreliable. DeCRIM (Ferraz et al., EMNLP 2024, arXiv:2410.06458) has the model decompose the instruction and a critic model decide where to rewrite. LexInstructEval (arXiv:2511.17561) evaluates with a formal grammar and a purely rule-based engine. Validate-and-re-ask in industry (Guardrails AI, Pydantic-style validators) is the same loop. There are also training approaches (RECAST, arXiv:2505.19030;

ImpRIF, arXiv:2602.21228; Conifer, among others) that build data from verifiable constraints or use them as reward for fine-tuning, and new benchmarks (CCR-Bench, arXiv:2603.07886; CCTU, arXiv:2603.15309). On model self-reports, *Self-Reports Are Not Verification* (Pan et al., NeurIPS 2026 workshop, arXiv:2609.00652) argues, in the setting of evolutionary search, that a model's statements about what it did should be treated as claims to be verified; the setting differs from ours.

**How this report differs** (stating only our own choices, without judging others'):

- **Reading does not go through a model.** Prior work mostly has a model decompose the instruction. We read with rules; the user's wording is not handed to a model to interpret, and we report the limit of doing so (Section 7.1).
- **The checking tool is not the only judge.** Besides the tool's verdicts, we collect blind labels of content quality and user-level acceptability, and report concrete patterns of "passed by the tool but not actually usable" (Sections 5 and 8).
- **Delivery strategies.** Besides whole-answer rewriting, we measure a local patch that changes only the failing part, and a conservative strategy that hands back the original answer when the repair fails.
- **Failures and costs are on the books.** Empty answers, answers made worse by rewriting, looping in the thinking segment and extra time are all reported as results.
- **Material.** Chinese prompts and small Qwen-family models.

Several observations in prior work agree with results we obtained independently: DVR also finds model self-checking close to chance, length constraints the hardest, and no gain from self-reflection.

This section is the result of several searches, not a systematic review.

## 3 Architecture (at the level of principle)

The governance layer (called the "external mechanical layer" in the earlier reports) has four steps, all outside the model:

1. **Reading.** Rules extract checkable requests from the user's words: length (upper and lower bounds, "about N"), number of items, format (such as JSON, table, list), paragraph structure, title, required and forbidden words, grammatical person, answer language, forbidden character classes. This step does not call a model.
2. **Generation.** The model answers as usual. For length and number of items, the governance layer can count on the model's behalf during generation and decide when generation may end (length control below); the model does not have to count.
3. **Checking.** An independent checker measures the answer item by item. The checker and the reading rules share no code.
4. **Repair and delivery.** When a request is not kept there are three strategies:
    - **Whole-answer retry**: tell the model which items were not kept and their target values, and ask it to rewrite once.
    - **Local patch**: handle only the failing part; fall back to whole-answer rewriting when the failing part cannot be located.
    - **Conservative**: try mechanical fixes and the local patch in turn, adopt the repaired answer only if everything is then kept, and otherwise hand back the original answer.

In this architecture the model does two things: write the answer, and revise the answer when instructed. It does not decide what the user asked for, and it does not decide whether it kept it.

**Terminology.** "Auditable" here means that when something goes wrong, one can find which step it was, change that step, and say clearly what happened. It does not mean nothing goes wrong.

**What is public.** This report describes principles (which signals are used, at which step a decision is made) and does not publish the rule tables, thresholds, templates or implementation. Measurement data, protocols and per-item results are published separately (Section 10.4).

## 4 Benefits: specification level

**Setup.** 329 prompts: the synthetic set S3 (120 prompts, Traditional Chinese) and the rule-only subset of ComplexBench (Wen et al., NeurIPS 2024; 209 prompts, Simplified Chinese). Four models: Qwen2.5-3B/7B/14B-Instruct and DeepSeek-R1-Distill-Qwen-7B, all loaded with 4-bit quantisation, greedy decoding. The metric is "all-pass": an answer keeps every checkable item the governance layer read (673 items over the 329 prompts). For 25 prompts (3 in S3, 22 in ComplexBench) no checkable item was read; they count as all-pass in every arm and are in the denominator. They do not affect the difference between two arms or the test, but they add 25 to every arm's all-pass count. Comparisons use a paired test: the exact two-sided binomial McNemar test on discordant pairs only, with no continuity correction; the tables list the discordant counts.

The table below has three columns of different nature (Section 8 explains where they come from). **Original checker** (v1) and **corrected checker** (v2, corrected and then frozen) re-score the same answers with a different checker. **Rerun after the length-control fix** is different: the counting used by the governance layer's length control was made consistent with the checker, the governed arms generated new answers, and these were scored with v2. That column is the system's performance after a fix, not a change of checker.

**Table 1. All-pass counts (329 prompts per cell): bare model → with the governance layer (whole-answer retry); in parentheses, "only the governed arm passes : only the bare model passes"**

| Model | Original checker (v1) | Corrected checker (v2) | Rerun after length-control fix (v2) |
|---|---|---|---|
| 3B | 160 → 248 (88:0) | 167 → 233 (66:0) | 167 → 259 (92:0) |
| 7B | 186 → 267 (81:0) | 209 → 263 (57:3) | 209 → 284 (75:0) |
| 14B | 192 → 256 (64:0) | 215 → 268 (54:1) | 215 → 281 (66:0) |
| 7B reasoning | 141 → 216 (75:0) | 170 → 217 (53:6) | 170 → 238 (68:0) |
| 7B (requests read are appended to the prompt first) | 209 → 267 (58:0) | 225 → 268 (45:2) | 225 → 289 (64:0) |
| 7B reasoning (same) | 171 → 233 (62:0) | 188 → 222 (39:5) | 188 → 240 (52:0) |

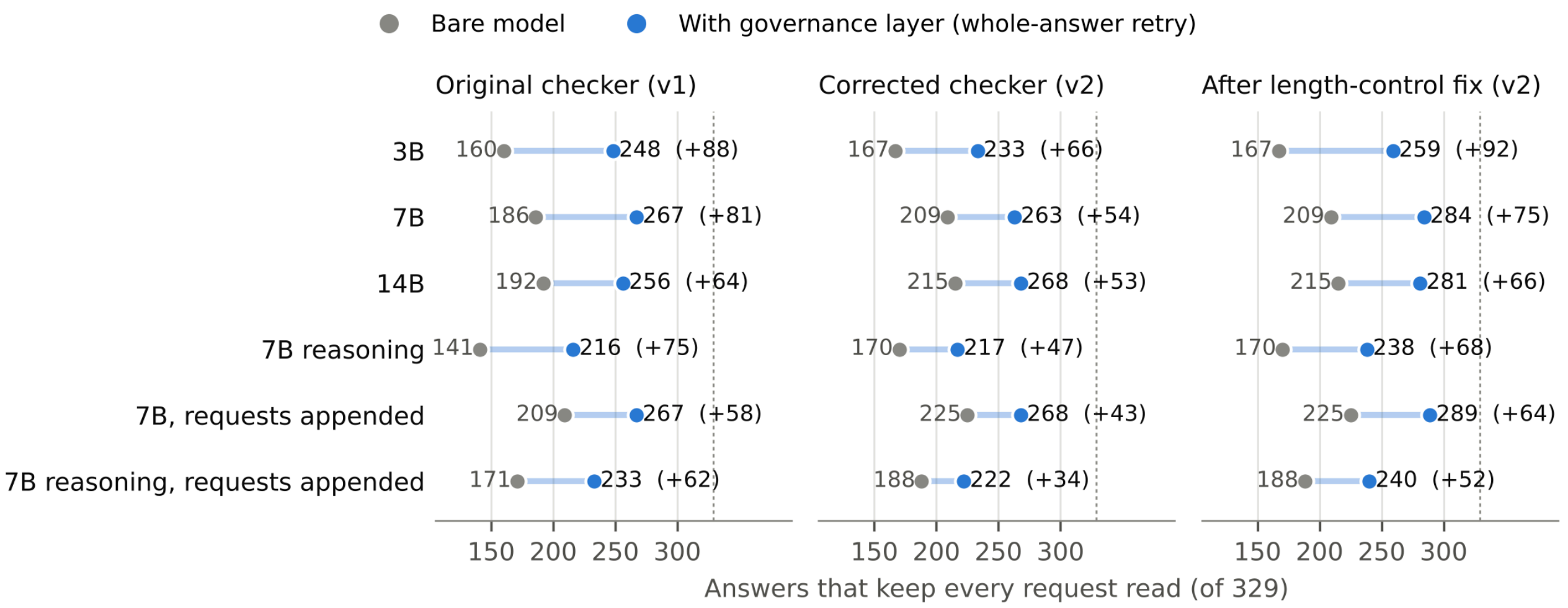

**Figure 1. The numbers of Table 1.** Each row is one model setting; the grey dot is the bare model and the blue dot is the same model with the governance layer (whole-answer retry), with the gain in parentheses. The dashed line marks 329, the number of prompts.

All eighteen comparisons have $p<0.0001$. Re-scoring the same answers (first two columns) does not change the conclusion. The gain is smaller under the corrected checker (+34 to +66) because v2 stopped counting punctuation toward length, which left the old length control inconsistent with the checker. After the length control was corrected and the arms rerun, the gain is +52 to +92.

**A generic-redo control.** In the control arm, when the bare model fails, it receives one generic redo message ("Please answer again and make sure you meet all the requirements in the prompt") without being told which item failed; the number of generations equals that of the governed arm. Under the corrected checker the discordant counts are 12:10 for 7B (net +2, $p=0.83$) and 16:10 for the reasoning model (net +6, $p=0.33$); under the same checker the governance layer has 57:3 and 53:6. That is, the one generic redo message we tested brought no significant gain. Comparing the two arms directly (only the governed arm passes : only the generic redo passes) gives 58:6 for 7B and 51:10 for the reasoning model (both $p<0.0001$): the complete governance layer outperforms this generic-redo control. This comparison cannot by itself identify which part of the governance layer makes the difference: besides "telling the model which item failed", the governed arm also has length control during generation. Nor does it show that other ways of retrying are ineffective; the control was run on two models only.

**Length alone.** Among the 90 repair-triggered prompts of Section 5, 72 have a length request. The number of answers in which every length item is met: bare model 12; conservative 40; whole-answer retry 55; local patch 60. Length here is counted on the "effective body": the text that remains after removing the model's pleasantries addressed to the user and exact duplicate sentences (see Section 8.3).

The numbers in this section say only whether the specification was followed. They are not used to infer how users feel.

## 5 Benefits: user level (exploratory)

**Question.** On prompts where the bare model failed a request, did the governance layer produce an answer the user can accept that the bare model could not? The question is whether the governance layer has this capability, so its three strategies are considered together. How each strategy performs on its own, and which one to choose, is the next step; see Appendix A.

**Method.** The sample was built as follows. For 3B and 14B separately, 60 prompts were drawn at random, stratified by the two prompt sets in proportion, from the prompts on which a repair was triggered (170 for 3B and 138 for 14B; 164 and 108 after removing prompts whose answers had already been used in other blind labelling). The checker and the length control were then corrected and the arms rerun. For 30 of these 120 prompts the four answers were character-for-character identical, leaving nothing to compare, and they were excluded: for 25, the bare answer already passes under the corrected checker and no repair is triggered; for 5, a repair was triggered and a rewrite was generated, but the rewrite was identical to the original answer. This leaves 90 prompts (49 for 3B, 41 for 14B), all of them repair-triggered. For each prompt the bare answer and the answers of the three governance strategies were shown side by side (identical texts shown once), with the source hidden, and the annotator selected the acceptable answers: every acceptable answer was to be selected, and selecting none was allowed. The scoring was fixed by the author after unblinding: a prompt where a governed answer (from any strategy) was selected and the bare answer was not counts as "governed"; a prompt where only the bare answer was selected counts as "bare"; prompts where both or neither were selected are excluded. The test is the exact two-sided binomial test.

The bare model has one answer because it can only produce one: decoding is greedy, and rerunning the same prompt gives the same result. A further control is letting the bare model answer again after a failure without telling it what failed (Section 4); all-pass rises by only 2 and 6 answers, neither significant.

The annotators are two AI models: the chat-side Claude model used in this research (CS below; on the research side) and GPT (an independent second annotator).

**Table 2. Governed : bare (3B and 14B combined, 90 prompts)**

| Annotator | As labelled | Selections short of the required length treated as not selected (exploratory) |
|---|---|---|
| CS | 49:7 ($p<0.0001$); 34 excluded | 43:3 ($p<0.0001$); 44 excluded |
| GPT | 18:12 ($p=0.36$); 60 excluded | 19:0 ($p<0.0001$); 71 excluded |

"Selections short of the required length treated as not selected" is a rule added after calibration (below); it only removes selections and nothing is re-selected. In all 12 prompts where GPT, as labelled, selected only the bare answer, the bare answer is short of the required length.

**Two sensitivity checks.** First, the right-hand column counts length on the effective body without a separate "deliverable" segmentation; keeping the original checker's deliverable segmentation as well gives 41:3 and 19:0, and the conclusion is unchanged. Second, 12 source prompts appear for both 3B and 14B among the 90 observations; keeping only one model's observation for each repeated prompt (78 observations) gives, as labelled, 43:3 (keeping 3B) and 44:5 (keeping 14B) for CS and 15:11 and 14:11 for GPT (still not significant), and after the length rule 38:0 and 38:3 for CS and 16:0 and 15:0 for GPT. In all four cells the direction and the significance status are unchanged. No multiple-comparison correction is applied to Table 2.

**An earlier single-choice round is not used.** The first version of the question was "choose the one answer you would use", with only one choice allowed. When labelling by hand, the author found this wording flawed: when two answers are both usable the annotator is forced to pick one, and the answer not picked is not thereby unacceptable, so the round cannot answer the question of this section. The question was changed to "select every acceptable answer", and this section uses only that round. The single-choice labels remain in the public data (CS 53:8, GPT 23:15); they point the same way and are not relied on here. CS's round was done after CS had seen the author's labels on 20 prompts and GPT's choices on those 20 prompts, so it is not fully blind; GPT's round does not have this problem.

**How to read it.** "49:7" means that in 49 prompts the governance layer produced an answer the annotator could accept and the bare model did not, and in 7 prompts the reverse. As labelled, the gap is clear for CS, and for GPT it points the same way but is not significant; after the length rule both are clear. The annotators were asked "which are acceptable". Many prompts are excluded, of two kinds:

- **Both sides acceptable**: the author's reading is that such prompts make no difference to the user and do not change which side he or she would pick; the user has nevertheless paid the time for the governance layer's checking and rewriting (Section 6).
- **Neither side good**: this is what the governance layer does not yet achieve.

**Human calibration of the labelling standard.** The author, acting as the user, labelled a stratified sample of 20 prompts in person, to align the standard with the two AI annotators; these labels are not in the results table. The calibration found that when a prompt has an explicit length request and no answer's body reaches it, the author always judged "none", whereas CS tended to pick "the closest one" (of the 7 prompts the author judged "none", CS agreed on only 2, GPT on 5); it also found that single choice forces a pick. The resulting rules: check whether the length is met first, then the content; an answer short of the required length is not acceptable; length is counted on the effective body; select every acceptable answer.

**Why exploratory.** There was no pre-registered decision rule; the scoring and the judging rules were both fixed along the way, after unblinding; the length-first rule is related to what the governance layer sets out to improve; CS is a model on the research side; the only human calibrator is the author, on 20 prompts; after calibration the two AI annotators did not re-label under the new rules, so the right-hand column of Table 2 is "pre-calibration labels plus mechanical removal under the calibrated rule". All 90 prompts are repair-triggered and do not represent prompts in general.

This section measures whether the governance layer "can produce it", not whether "that is the answer delivered every time". For each prompt the governance layer has up to three different answers that can be selected and the bare model has one; this is inherent in asking about the three strategies together, and in deployment only one answer would be delivered. Results for each strategy alone against the bare model are in Appendix A: no single strategy consistently beats the bare model, and choosing the right one of the three is future work.

## 6 Costs

**Time.** Reading and checking take 0.006 to 0.035 seconds per prompt, which is negligible. Almost all the extra time is the one additional generation for the rewrite.

**Table 3. Time cost of whole-answer retry (measured; same machine, 4-bit, one prompt at a time)**

| Model | Bare, seconds per answer | Extra seconds per answer | Extra share | Marginal seconds per additional all-pass answer (v1 / v2) |
|---|---|---|---|---|
| 3B | 18.6 | 12.3 | 66% | 46 / 61 |
| 7B | 14.6 | 7.9 | 54% | 32 / 48 |
| 14B | 24.9 | 13.2 | 53% | 68 / 82 |
| 7B reasoning | 52.5 | 22.3 | 43% | 98 / 156 |

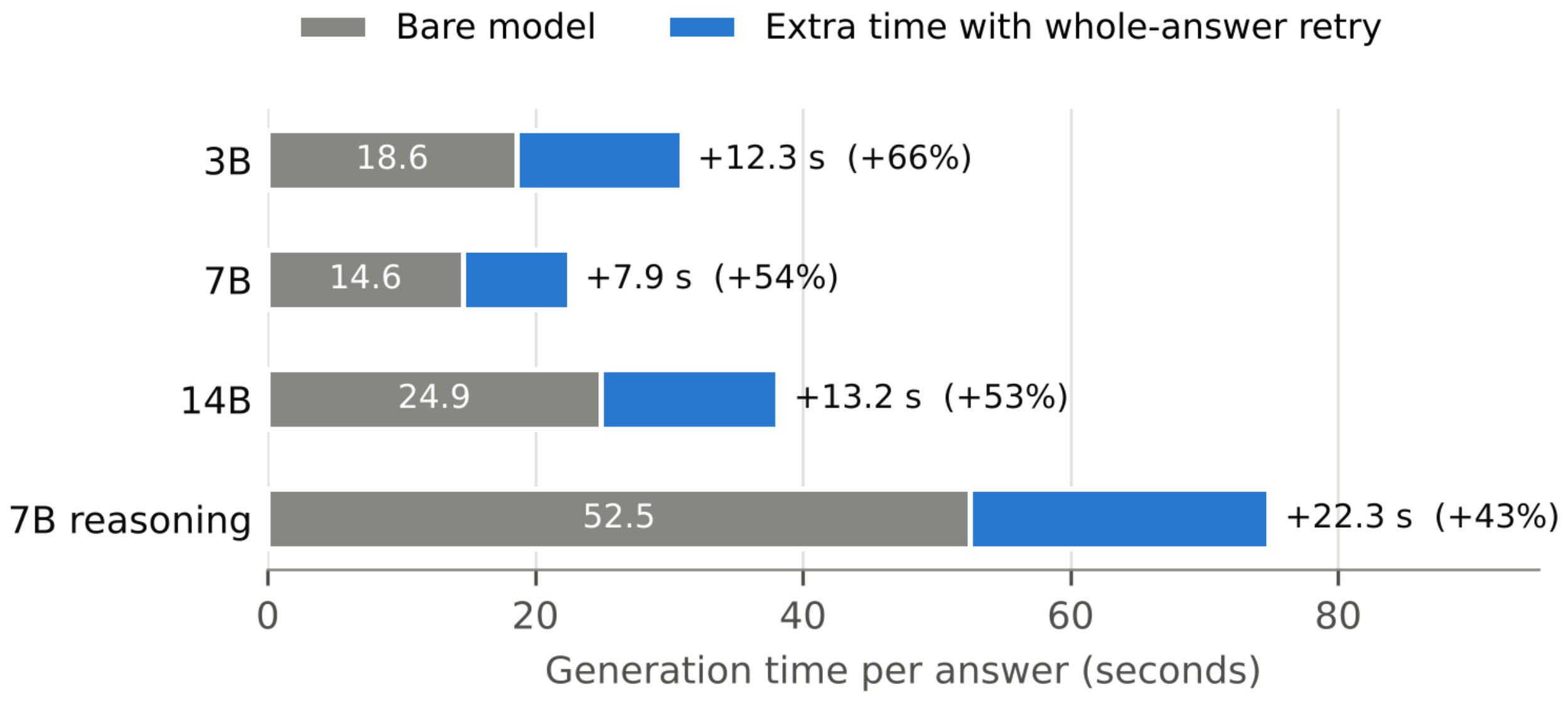

**Figure 2. The times of Table 3.** Generation time per answer for the bare model (grey) and the extra time added by whole-answer retry (blue).

Table 3 measures whole-answer retry before the length-control fix; the marginal seconds correspond to the gains under the original and the corrected checker. Time for the rerun after the length-control fix was not measured separately. Time for the local patch was not measured; only the number of calls was recorded (for the four models, 114 to 162 answers handled with 83 to 131 calls each). The conservative strategy adopts the product of the local patch and makes no model calls beyond it, so its cost is not lower than that of the local patch. Time varies with hardware, quantisation and batching and is meaningful only for comparison under the same conditions.

**Whole-answer rewriting damages content.** Blind labels of content quality (usable / truncated / content worse) show that rewriting often turns a usable answer into an unusable one:

- 7B and the reasoning model, 60 answers in total: 57 usable for the bare model, 36 for whole-answer retry, 45 for the local patch. Bare → whole-answer retry: 23 worse, 2 better ($p<0.001$); bare → local patch: 13 worse, 1 better ($p=0.002$).
- 14B, 30 answers, two annotators: bare → governed, worse versus better is 14:4 (CS, $p=0.031$) and 10:1 (GPT, $p=0.012$).

**Table 4. Effect of repair on "usable" (60 answers per row; left: the original answer is usable and the delivered version is not, with the share of originally usable answers in parentheses; right: the original answer is not usable and the delivered version is)**

| Model / annotator | Original usable | Broken by whole-answer retry | Broken by local patch | Broken by conservative | Original not usable | Rescued by whole-answer retry | Rescued by local patch | Rescued by conservative |
|---|---|---|---|---|---|---|---|---|
| 3B / CS | 42 | 22 (52%) | 19 (45%) | 9 (21%) | 18 | 2 | 2 | 0 |
| 14B / CS | 49 | 33 (67%) | 26 (53%) | 12 (24%) | 11 | 2 | 2 | 0 |
| 3B / GPT | 23 | 18 (78%) | 15 (65%) | 6 (26%) | 37 | 1 | 2 | 1 |
| 14B / GPT | 32 | 20 (63%) | 13 (41%) | 9 (28%) | 28 | 0 | 3 | 2 |

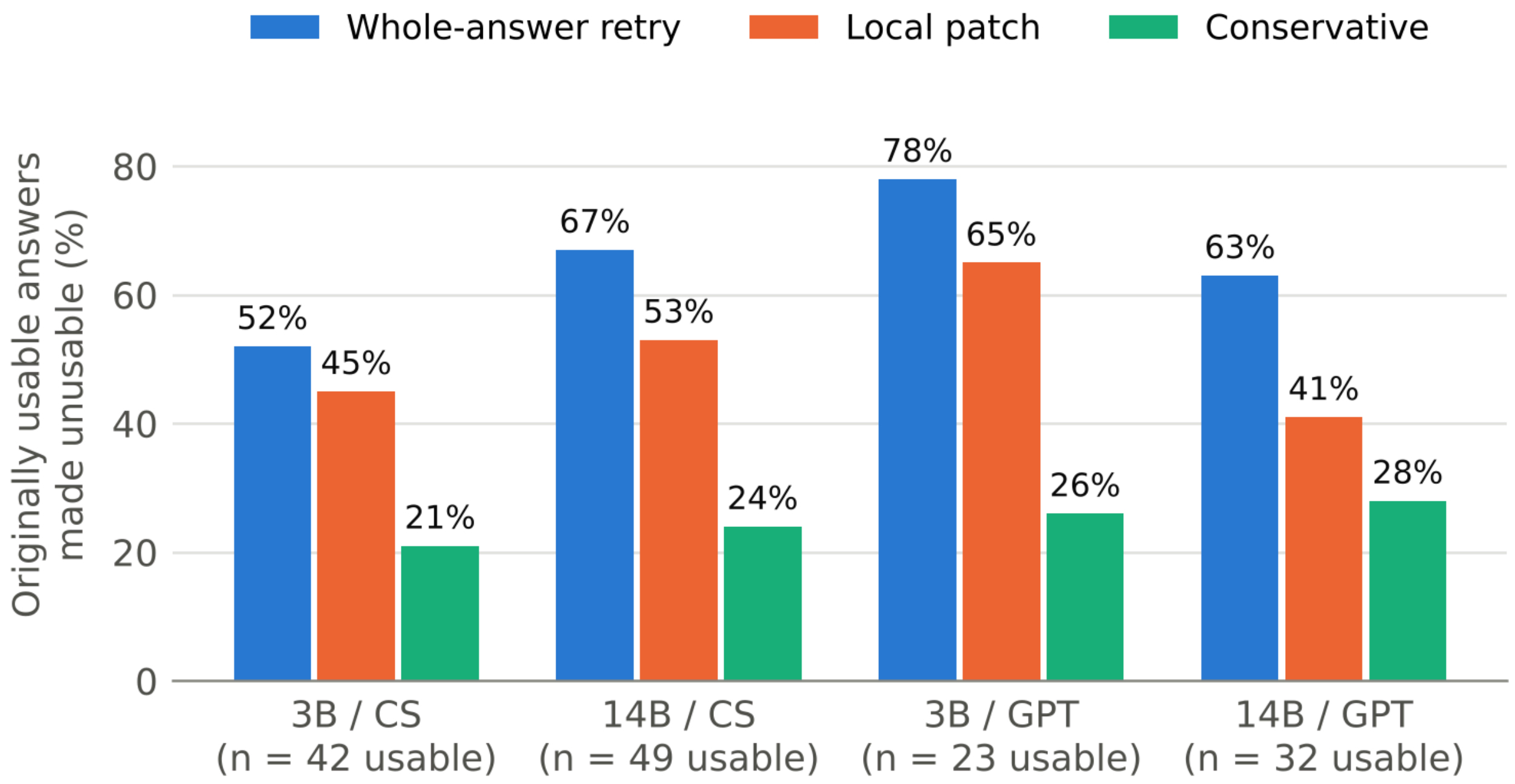

**Figure 3. The left half of Table 4.** Share of originally usable answers that the delivered version made unusable, by strategy, for each model and annotator. The rescued counts (right half of Table 4) are 0 to 3 per cell and are not drawn.

Each row is a sample of 60 answers, all from repair-triggered prompts; the proportions cannot be extrapolated to prompts in general. On the "usable" dimension, far fewer answers get better than get worse: the repair makes the answer keep the specification, but content quality does not improve as a result and often declines.

In all four rows the conservative strategy breaks the fewest answers. It adopts a repair only when everything is then kept, which is a possible reason; we did not verify this separately. Agreement between the two annotators is moderate (303 answers for 3B and 14B combined, $\kappa \approx 0.49$), and GPT is stricter overall; the direction is the same for both.

**Empty answers.** Across all runs there are 3 truly empty answers (0 characters of body), all from the reasoning model (Section 7.2). A further 44 answers were close to empty; on in-spection, 11 were due to misreading by the reading step, 14 to mutually exclusive requests, and 19 to prompts that call for a short answer in the first place.

# 7 Failures

## 7.1 The limit of reading

The reading exams were written by the chat-side model, with sentences in phrasings the governance layer had not seen; a second annotator labelled them independently to estimate "the upper bound on which two readers agree" (the ceiling).

**Table 5. Reading against the ceiling** (thresholds were set in advance; what is compared is the "against ceiling" column, not the raw proportion)

| Exam (sentences) | Metric | Governance layer | Ceiling | Against ceiling | Threshold (against ceiling) | Verdict |
|---|---|---|---|---|---|---|
| Exam 3 (125) | Newly phrased requests that were read | 22/35 (63%) | ≈90% | ≈70% | ≥85% | Not met |
| Exam 3 | Trap sentences not taken as requests | 34/40 (85%) | ≈100% | ≈85% | ≥90% | Not met |
| Exam 3 | Conflicting requests detected | 13/15 (87%); false alarms 0/10 | ≈93% | ≈94% | ≥90%, 0 false alarms | Met |
| Exam 4 (115) | Newly phrased requests that were read | 32/38 (84%) | ≈100% | 84% | ≥85% | Missed by 1 item |
| Exam 4 | Trap sentences not taken as requests | 34/40 (85%) | 100% | 85% | ≥90% | Not met |
| Exam 4 | Conflicting requests detected | 8/10 (80%); false alarms 0/10 | 100% | 80% | ≥90%, 0 false alarms | Not met |

The denominator of each metric is the number of sentences of that kind in the exam, not the number of sentences in the whole exam. The ceiling is the proportion on which the two annotators agree, and it serves as the reference for the thresholds: a threshold is set relative to the measured ceiling, not relative to 100%. It is an estimate of "how much two readers can read in agreement", an operational reference chosen in this research for setting thresholds. It is not a theoretical upper bound on reading ability, and it does not mean a system cannot exceed it; agreement between the two annotators also does not guarantee that both are right.

Only one of the six cells meets its threshold. For newly phrased requests the result against the ceiling is about 70% (Exam 3) and 84% (Exam 4); trap sentences are left alone 85% of the time on both exams. Reading errors have two consequences. A request that is missed is checked by no one, so for that item the situation is the same as having no governance layer. Taking something that is not a request (something someone else said, a sentence describing something else) as a request makes the governance layer tell the model to rewrite for a request the user never made. We judge the latter to be the more serious, because it changes an answer that had no problem; this is a design judgement, and the harm caused by each kind of error was not measured separately. The false-trigger rate on trap sentences is 15% on both Exam 3 and Exam 4. There is also a "right result, wrong reason" pattern (scope, conditionals or counting units read wrongly), which is not fixed in this version.

### 7.2 Mutually exclusive or unsatisfiable requests

All 3 truly empty answers come from the reasoning model looping in its thinking segment until the generation budget is used up; the items that triggered them are requests that are hard to satisfy, conflict with one another, or were misread. In these 3 cases, sending back a request that cannot be met left the reasoning model spinning without delivering anything. The cases are few and there is no control; we record this only as an observed failure pattern and draw no general conclusion. In the 2 cases where a draft appeared in the thinking segment, the draft would not have passed the check either. Conflicting requests can be detected in advance (Table 5), but the step "when detected, ask the user first instead of sending back" is not yet connected in this version.

### 7.3 The model's self-reports are unreliable

**Says fixed, not actually fixed.** After the governance layer marked what was wrong and what the target values were, the 7B model was asked to change only those and report item by item. Of the 25 items it reported as "fixed", only 7 pass on re-checking; 18 (72%) were not fixed. In 13 of the 27 answers it gave only the report and no answer, and in 9 it broke items that had passed before. In the end this approach was not adopted for any answer. The 25 items come from the 27 answers of one 7B repair experiment; items within the same answer are not independent, and they cannot be treated as 25 independent behaviours.

**Unprompted claims of meeting the length.** Among the 233 answers of Section 5, a program identified 29 that claim, unprompted, to meet the length or that state a count; each was compared with its body: 12 true, 16 false, 1 not checkable. Two points about how they were checked: when a claim refers to only part of the answer, it is still judged against the length item for the whole answer, and that part is not checked separately; for stated counts, a tolerance of plus or minus ten per cent is allowed. No answer says outright that it falls short of the length. Example: the answer says "the content above totals more than the required 501 characters and fully meets your needs", and its body has 438 characters.

These two findings are the direct reason this work places checking outside the model: the model saying it did something cannot be taken as evidence that it did.

## 8 Calibrating the instrument

The measuring tool itself makes mistakes. This section reports how the checker and the length count were examined and corrected, and which conclusions changed afterwards.

### 8.1 Auditing the checker against the truth

From the answers of three models, 214 "item–verdict" pairs were sampled, and two AI auditors (CS and GPT) judged independently whether the checker's verdict was right. The two audits agree on "checker right or wrong" for 187 of 214 (87.4%).

Main findings (counts below are the first auditor's, CS): on requests that require parsing the structure of the answer first (number of items, structure, format, title containing a word), of the 94 "violated" verdicts of the original checker, 49 were measurement errors by the checker, 15 were cases where the reading did not match the prompt's meaning, and only 30 were right. Typical measurement errors: numbered lists not recognised as lists; elements of a JSON array not counted as items; a first line not recognised as a title; the same word in Traditional and Simplified Chinese treated as different words. Surface-level requests (length, forbidden characters and the like) are largely reliable; in a sample of 60 "kept" ver-dicts no measurement error was found. Pairs judged among the 214 as a reading mismatch or as undecidable are problems of the reading layer and are not counted for or against the checker.

## 8.2 Corrections, conventions and freezing

The disagreements between the two audits concentrate on a few conventions, on which the author ruled at the level of principle (not item by item): answer language is judged on the delivered content, not on the pleasantries before and after it; Traditional and Simplified forms are treated as the same character; pure whitespace in JSON may be normalised before parsing; "right result, wrong reason" counts as a checker defect; length does not count punctuation and follows the word-processor convention (each Chinese character counts one, each run of consecutive Latin letters or digits counts one); English words are matched as whole words, with inflections. With the principles given, the second auditor re-audited the 19 items still in disagreement (of 27 items originally in disagreement, the first auditor changed 8 after reading the other's reasons) and agreed with the first auditor on all of them. This shows that consistent verdicts can be reached under the author's principles; it is not additional independent evidence.

The checker was corrected accordingly into v2. On the 176 scorable items, v1 is right on 114 and v2 on 173; within these, on the 77 structural items the count rises from 25 to 77. Note that these 176 items are the very set used to find the problems and drive the corrections: 173/176 is performance on the correction set, not independent validation, and there is no separate validation set in this work. In addition, 86 items whose verdict changed under v2 were sampled for a flip audit; after merging the two audits one more fix was made (explanatory sentences that talk about the constraint itself are excluded from word matching), and the checker was then signed off and frozen. Problems found after freezing are left for the next version and the checker is not changed.

## 8.3 Length

Three things were corrected:

- **How the checker counts**: changed from counting punctuation to not counting punctuation (the convention above).
- **How the length control counts**: the governance layer's counting during generation originally included punctuation, which did not match how users count; after it was changed to the same convention the governed arms were rerun (third column of Table 1).
- **What is counted**: human calibration found that the model's pleasantries addressed to the user and copy-pasted duplicate sentences were being counted toward length, so that answers "meeting the length" actually had too little body. The effective body was defined accordingly and is used for the right-hand column of Table 2 and for the length results in Section 4. Counted with the frozen v2 convention, the length results are bare 11, whole-answer retry 62, local patch 64, conservative 47; counted on the effective body they are 12, 55, 60 and 40.

## 8.4 Which conclusions changed after the corrections

- **The gain from the governance layer**: re-scoring the same answers, all six comparisons remain significant and the conclusion is unchanged; likewise after the length control was corrected and the arms rerun (Table 1).

- **Local patch versus whole-answer retry**: downgraded. The main metric is "usable and all-pass", with 3B and 14B combined. The decision rule set in advance was: CS's paired test has $p<0.05$ with positive direction, and GPT's direction is positive:

| Checker | CS (only local patch wins : only whole-answer retry wins) | p | GPT | p | Verdict under the pre-set rule |
|---|---|---|---|---|---|
| Original checker, confirmatory | 17:6 | 0.035 | 14:1 | 0.001 | Holds |
| Corrected checker, sensitivity | 15:6 | 0.078 | 14:2 | 0.004 | Does not hold |
| Re-judged after the length-control fix (exploratory) | 8:5 | 0.58 | 8:0 | 0.008 | Does not hold |

Our wording is: same direction, unstable significance, with no conclusion drawn from the p-value of any single row. We neither say on this basis that the local patch is better nor say that the difference has disappeared.

- **Conservative versus bare**: holds under every checker (original 31:1; corrected 13:2; after the length-control fix 18:0). But this comparison is structural: these are prompts where the bare model failed and a repair was triggered, so the bare model has almost no chance to score on "usable and all-pass". The 1 to 2 prompts won by the bare model alone arise from a difference between the versions used at trigger time and at scoring time: in one, the reading version used in the run read a request and triggered, and the newer reading version used for scoring no longer reads that request; in the other, the length was out of range under the original checker and falls within range once punctuation is not counted. This comparison therefore shows only how many usable and all-pass answers the conservative strategy can deliver on triggered prompts; it is not a comparison of merit.

## 9 Discussion

**A range (the author's reading).** With the model and the decoding settings held fixed, there is nothing to adjust in the bare model; its result on these prompts is the number observed. The governance layer's result can change when the rules are revised, and for each prompt the reason it fell short can be found. Take length alone: of 72 prompts, the bare model meets the length in 12, the three governance strategies in 40 to 60, and the full score is 72. All three numbers are observations under the settings of this report: 12 is not an upper bound for the bare model under every setting (a different prompt or a different model may change it; that is not the work of this research); 40 to 60 is not a guaranteed lower bound; 72 is the ideal full score, not a number shown to be attainable. The statement is also an aggregate one and does not mean the governance layer is no worse on every prompt: Table 2 still contains prompts where only the bare answer was selected (7 for CS, 12 for GPT).

**Division of labour.** Understanding what the user did not say outright, and how well the content itself is written, are mainly the model's ability. The governance layer is responsible for reading correctly what the user said outright and measuring the answer correctly; telling the user truthfully what was not achieved is for a later stage (Section 11). Our data show that reading and measuring can be done without the model that generates the answer, and also show where the limit of each lies. But the governance layer does not only read and measure: it tells the model to rewrite, and rewriting damages content (Section 6). That damage is the governance layer's responsibility and cannot be attributed to the model.

# 10 Limitations and statements

## 10.1 Conditions under which the conclusions hold

The conclusions hold only under these conditions:

1. **This set of material.** The synthetic set S3 (120 prompts, Traditional Chinese; 72 built from original material by the chat-side model plus signed-off constraint templates, 48 written originally by the execution side on topics taken from popular threads of an online forum, with no text copied) and the rule-only subset of ComplexBench (209 prompts, Simplified Chinese; those of the original 1,150 prompts whose constraints can all be checked by rules). The constraint types are the explicit requests listed in Section 3.
2. **One model family.** Qwen2.5-3B/7B/14B-Instruct, and DeepSeek-R1-Distill-Qwen-7B, which is built on Qwen2.5-Math-7B.
3. **Model sizes up to 14B.** All four models are loaded with 4-bit quantisation (NF4, double quantisation), not at full precision.

We make no inference about other model families, larger models, other languages or other task types.

**Single turn only.** Every result in this report comes from a single-turn setting: one user message and one answer. Requests that carry over from an earlier message, or that are withdrawn or changed in a later message, were not tested; the reading step looks only at the current message. Real conversations are multi-turn, and we make no inference about them.

## 10.2 Measurement limitations

- The annotators for content quality, for the checker audit and for the user level are all AI (CS and GPT). CS is a model on the research side and has a stake. The human (the author) ruled on principles and calibrated on 20 prompts, and did not spot-check item by item.
- Agreement between the two annotators is moderate (quality $\kappa \approx 0.49$; 87% binary agreement in the checker audit).
- The 90 user-level prompts are all repair-triggered; the judging rules were fixed along the way; no re-labelling was done after calibration.
- The reading exams were written by the research side.
- The four models were tested in one setting only: 4-bit quantisation and greedy decoding.
- Only the subset of ComplexBench whose constraints can all be checked by rules was used.
- The checker corrections have no independent validation set (Section 8.2).
- The same prompt appears repeatedly in comparisons across models, and S3 prompts share constraint templates, so prompts are not fully independent. The only sensitivity check is the one for Table 2 in which each repeated prompt is kept for one model only (Section 5); shared templates are not addressed, and no multiple-comparison correction is applied. The exploratory comparisons in particular should be read conservatively.
- The early quality labels did not systematically verify facts in the answers.
- The prompts are synthetic or adapted from a public benchmark, not real usage traffic. The earlier reports measured an effect close to zero on samples of real Chinese conversations, mainly because requests with explicit constraints are rare in those samples.

## 10.3 Disclosure of AI collaboration

The author is responsible for the research design, the rulings and the public text. Implementation and batch runs were carried out with Claude Code; Claude (chat side) took part in writing exams, analysis, labelling and drafting; second labelling and review were done with GPT. The author is responsible for the whole text.

## 10.4 Reproducibility

There are two kinds of reproducibility, and this report achieves only the first:

- **Recomputable.** Published: the prompts, each arm's answers, per-answer and per-item verdicts, labelling and audit records (including the unblinded keys of the blind tests), the principles ruled on, the checker (a standalone version without the governance layer), the data pack and the programs that produced it, and model and environment information. What an outside party can do falls into three tiers. First, from the published answers and item list, recompute every item's verdict with the published checker, and the tables derived from the verdicts (Table 1, Section 8.4). Second, from the published labels and unblinding keys, rebuild the label-based statistics (the "as labelled" column of Table 2, the times in Table 3). Third, some results can only be checked against existing summaries and cannot yet be rebuilt end to end from the raw answers: the length-removal column of Table 2, effective-body lengths, and the self-report checks of Section 7.3; the programs for these depend on internal files that are not published. The item list comes from the governance layer's reading; an outside party can only re-check against the published list and cannot re-extract items from the prompts.
- **Not independently re-runnable.** The reading rule tables, thresholds, templates, and the repair and length-control implementations are not published. An outside party cannot regenerate the governed answers or rerun the same method on new data. This is a substantive limitation of this report. The implementation can be requested from the author and is subject to the author's review.
- **One piece of evidence is published only in part.** Of the item-level evidence for Section 7.3, two files contain the original text of retry messages and are outside the published scope; the remaining assembly and blind-validation files are published, together with the numbers given here. In the published answers, wording the model copied from the governance layer is masked with a placeholder string (71 places, listed in `data/v3/answers/REDACTIONS.csv`); verdicts were computed on the text before masking, so recomputing these answers from outside may differ from the published verdicts.

Location: `https://github.com/blackwing04/Logical-AI-OS`, tag `v3.0-data` (commit `6d9226c6a702989f54fb8a28775b8a93d1cc3b0d`). Data are in `data/v3/`, the checker and the programs that produced the data pack in `eval/v3/`, per-item verdicts and the data pack in `results/v3/`, and the SHA-256 of each file in `manifest/v3_manifest.csv`.

# 11 Next stages

Directions only, with no unfinished numbers reported: check on the effective body and deliver truthfully (the governance layer states by how much a request was missed); an evaluation protocol that measures the two levels separately; re-validation on other model families;

pre-registered confirmation on new, unseen prompts; asking the user first when requests are mutually exclusive; a pre-execution gate for agent settings; giving the model a reference to check against.

## Appendix A. Each of the three governance strategies against the bare model (internal diagnosis)

Section 5 establishes that the governance layer can produce acceptable answers the bare model cannot. This table goes on to ask which of the three strategies is better; it points the way for the next improvement of the governance layer and is not to be read together with "governed versus bare" in the main text. Scoring: each time only one strategy's answer and the bare answer are considered; prompts where the two texts are identical are excluded entirely. 3B and 14B combined, exact two-sided binomial test. The three strategies share the same prompts and are not independent; no multiple-comparison correction is applied.

| Basis | Annotator | Whole-answer retry : bare | Local patch : bare | Conservative : bare |
|---|---|---|---|---|
| As labelled | CS | 30:26 (0.69) | 33:25 (0.36) | 16:8 (0.15) |
| As labelled | GPT | 16:22 (0.42) | 10:18 (0.18) | 4:13 (0.049) |
| Short of length treated as not selected | CS | 25:18 (0.36) | 29:17 (0.10) | 15:4 (0.019) |
| Short of length treated as not selected | GPT | 16:7 (0.093) | 11:4 (0.12) | 5:1 (0.22) |
| Prompts excluded because the text equals the bare answer | | 1 | 0 | 46 |

What can be seen: no single strategy consistently beats the bare model; the difference becomes clear only when they are taken together (Table 2). The conservative strategy delivers the original answer in 46 prompts, which shows that its adoption condition is strict. As labelled, GPT leans toward the bare model in the conservative column (the local-patch column points the same way but is not significant); this disappears after the length rule is applied.

## Appendix B. Operational definitions used in measurement

These are the definitions adopted in this research; we do not claim they are how every user counts. Threshold numbers in the text are indicative settings.

- **Length**: punctuation, symbols, emoji, markup syntax and whitespace are not counted. Letters and digits of non-Latin scripts (Chinese characters, kana and so on) count one each; each run of consecutive Latin letters or digits counts one. For items where the prompt asks for a count in English words (unit "words"), English words are counted instead.
- **"About N"**: within 20% above or below N, inclusive (for example, "about 200 characters" is 160 to 240).
- **Effective body**: first remove what the model says to the user: announcement-style lead-ins at the start, pleasantries and hand-over remarks at the end, and sentences in which it states its own length or claims to meet the requirements (erring on the side of removing too little rather than removing wrongly); when the prompt itself asks for comments, reasons or explanation, closing comments are kept. Then remove duplicate sentences: sentences identical to earlier text after normalisation (unifying Traditional and Simplified forms, removing whitespace, list markers at the start of the sentence and bold markup), keeping the first occurrence. Sentences shorter than 5 counted characters are not treated as duplicates; duplicate sentences shorter than 15 counted characters are removed only when they fall inside a block of two or more consecutive duplicate sentences; duplicates are not removed for prompts asking for song lyrics.

- **Answer language**: judged on the delivered content. Only when the request is to answer in English, consecutive lines at the head and tail are removed if they are mainly Chinese, no longer than 60 characters, and of an opening or closing pattern; Chinese inside the body is still counted.
- **Traditional and Simplified**: for required words, words required in the title and forbidden words, both sides are converted character by character to Traditional Chinese before matching.
- **English words**: matched on word boundaries, case-insensitively, including noun and verb inflections (not adjective comparatives); multi-word phrases are not inflected and are matched as a whole string.
- **Explanatory sentences about required and forbidden words**: before matching, sentences that "mention the word in quotation marks, immediately next to constraint wording" are excluded, sentence by sentence (for example: "This reply avoids using the word 'responsibility'").
- **JSON**: only when parsing the original text fails are pure-whitespace characters replaced with ordinary spaces before parsing again; other invisible control characters are judged invalid.
- **Number of items**: the number of list lines plus the number of table data rows, then the maximum of that sum, the number of elements of a JSON array, and the longest run of consecutive numbering within one line; when all are zero, the number of blocks separated by blank lines. The verdict is "at least N": more than N also passes.
- **Title**: a title is present if any of the following holds: a heading line in markup syntax; a short line ending in a colon; a line beginning with "標題", "題目" or "Title"; a whole line marked with 【】, 《》, bold or italics; a section line with a Chinese ordinal; a first line of at most 40 characters that does not end with sentence-final punctuation or a comma and is not a lead-in.

The "at least N" rule for the number of items means that when the user asks for "five points" and the answer gives seven, our metric counts it as kept.